\documentclass[letterpaper,10 pt,journal,twoside,nofonttune]{IEEEtran}

\usepackage{cite}
\usepackage{amsmath,amssymb,amsfonts}
\usepackage{algorithmic}
\usepackage{graphicx}
\usepackage{textcomp}
\usepackage{xcolor}
    
\usepackage{algorithm}
\usepackage{multirow}
\usepackage{amsmath}
\usepackage{amsthm}
\usepackage{bm}
\usepackage{siunitx}

\usepackage{array}
\usepackage{tabularx}
\usepackage{booktabs}
\usepackage[super]{nth}

\usepackage[caption=false,font=footnotesize]{subfig}

\usepackage{url}
\usepackage[colorlinks,linkcolor=black,anchorcolor=black,citecolor=black,urlcolor=black]{hyperref}

\usepackage{threeparttable}
\usepackage{microtype} 

\usepackage{soul}
\usepackage{color, xcolor}

\soulregister{\cite}7   
\soulregister{\citep}7  
\soulregister{\citet}7  
\soulregister{\ref}7    
\soulregister{\pageref}7 
\soulregister{\textit}7 
\sethlcolor{yellow}

\usepackage{nomencl}
\makenomenclature

\begin{document}

\title{ToMPC: Task-oriented Model Predictive Control via ADMM for Safe Robotic Manipulation}

\author{Xinyu~Jia,
    Wenxin~Wang,
    Jun~Yang,
    Yongping~Pan,~\IEEEmembership{Senior~Member,~IEEE}, \\%
    and~Haoyong~Yu,~\IEEEmembership{Senior~Member,~IEEE}%
    \thanks{Received 24 January 2025; accepted 28 May 2025. This article was recommended for publication by Associate Editor M. Benallegue and Editor O. Stasse upon evaluation of the reviewers’ comments. This work was supported by the Singapore Ministry of Education Academic Research Fund (AcRF) Tier 2 under Grant T2EP50222-0009. (\textit{Corresponding author: Haoyong Yu}.)}
    \thanks{Xinyu Jia, Wenxin Wang, Jun Yang, and Haoyong Yu are with the Department of Biomedical Engineering, National University of Singapore, Singapore 117583 (e-mail: xinyu.jia@u.nus.edu; wenxin.wang@u.nus.edu; bieyang@nus.edu.sg; bieyhy@nus.edu.sg).}
    \thanks{Yongping Pan is with the Peng Cheng Laboratory, Shenzhen 518057, China, and also the School of Electrical and Electronic Engineering, Nanyang Technological University, Singapore 639798 (e-mail: eee-yppan@ntu.edu.sg).}
    \thanks{Digital Object Identifier (DOI): see top of this page.}
}

\markboth{IEEE Robotics and Automation Letters. Preprint Version. Accepted June, 2025}
{Jia \MakeLowercase{\textit{et al.}}: ToMPC: Task-oriented Model Predictive Control via ADMM for Safe Robotic Manipulation} 

\maketitle

\begin{abstract}
    This paper proposes a task-oriented model predictive control (ToMPC) framework for safe and efficient robotic manipulation in open workspaces. The framework unifies collision-free motion and robot-environment interaction to address diverse scenarios. Additionally, it introduces task-oriented obstacle avoidance that leverages kinematic redundancy to enhance manipulation efficiency in obstructed environments. This complex optimization problem is solved by the alternating direction method of multipliers (ADMM), which decomposes the problem into two subproblems tackled by differential dynamic programming (DDP) and quadratic programming (QP), respectively. The effectiveness of this approach is validated in simulation and hardware experiments on a Franka Panda robotic manipulator. Results demonstrate that the framework can plan motion and/or force trajectories in real time, maximize the manipulation range while avoiding obstacles, and strictly adhere to safety-related hard constraints.
\end{abstract}


\section{Introduction}
\label{intro}

\IEEEPARstart{I}{ncreasing} demands in open workspaces require robotic manipulators to possess versatile motion and interaction capabilities \cite{Sami_agent_2023}. Beyond performing general pick-and-place tasks \cite{Jia_icra_2024}, robots must address safety concerns posed by the environment, such as avoiding collisions with obstacles \cite{huang_apf_2023}, or ensuring controllable external contacts \cite{Kim_contactMPC_2024}. In motion planning, random sampling-based approaches can rapidly generate collision-free paths; however, they often overlook task execution requirements and face challenges in physical interactions \cite{noreen_rrt_2016}. By contrast, optimization-based methods provide a convenient framework to incorporate diverse objectives and enforce complex constraints \cite{Sami_agent_2023}. Among optimization methods, model predictive control (MPC) stands out for its capability of online replanning \cite{Carlos_ddp_2023}, enabling adaptability in dynamic and uncertain environments. Therefore, this work will leverage the MPC approach to improve robot safety and manipulation behavior.

In the field of MPC applications, there are extensive studies for safe robotic manipulation. For instance, adding velocity damping terms in the cost function allows manipulators to stably pass through singular configurations \cite{Lee_mpc_2023}. Additionally, extending the prediction horizon has been shown to reduce velocity constraint saturation \cite{kleff_high_2021}. For obstacles in the workspace, enforcing distance constraints can prevent the robot from collisions \cite{hu_thread_2021}. While the MPC handles position and velocity objectives or constraints effectively, it may struggle with force-based manipulation if only using kinematic information. In \cite{admittance_2016}, the authors design a MPC-based admittance controller to maintain a desired contact wrench. Alternatively, dynamics-based formulations can directly overcome this issue, albeit at the expense of increased computational burden. In \cite{impedance_2020}, a model predictive impedance control is developed with interaction dynamics. Later, simplified robot dynamics are introduced to support a sequence of manipulation tasks \cite{mpic_2023}. In recent years, full-body dynamics are integrated into contact problems \cite{zhou_deform_2023}. However, few studies consider obstacle avoidance while applying contact force. The manipulation quality is also often ignored in avoidance. The gap is probably due to the increased complexity of the problem, posing challenges for real-time implementation.

\begin{figure}[!t]
    \centering
    \includegraphics[width=0.45\textwidth]{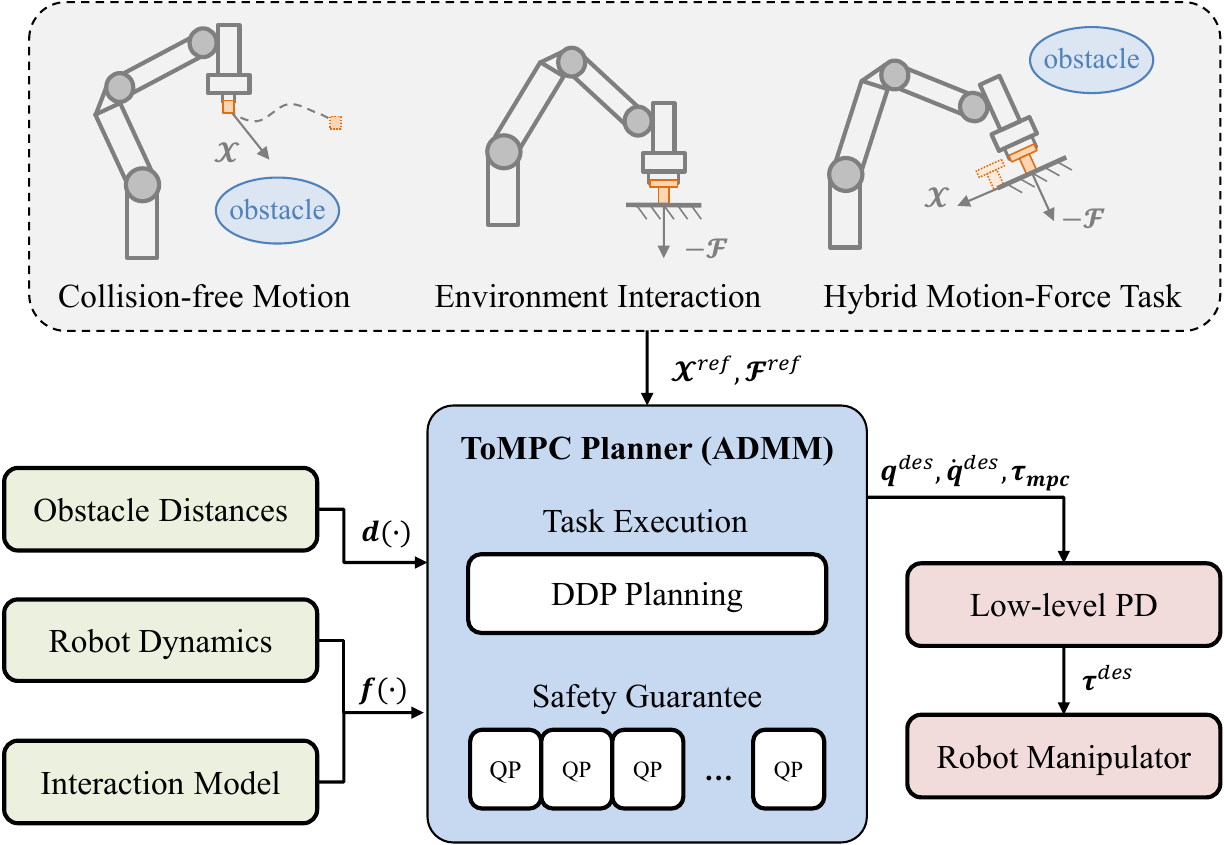}
    \caption[]{A schematic diagram depicting the the proposed ToMPC planner, where the ADMM solves the optimization problem in a distributed fashion. }
    \label{fig:framework}
    \vspace{-0.3cm}
\end{figure}

The computational efficiency of MPC remains a concern in robotics. To address this challenge, a common approach is to linearize the system model and transcribe the MPC into a quadratic programming (QP) problem, where convexification of the QP enables very high replanning frequencies \cite{Lee_mpc_2023}. Other acceleration techniques include exploiting multiple threads of processors \cite{hu_thread_2021} or reformulating the problem using multiple-shooting method \cite{jia_multi-layered_2024}. Recently, differential dynamic programming (DDP) \cite{zhou_deform_2023} and its first-order variants \cite{eth_mpc_2022} are widely adopted to solve nonlinear optimization problems. The backward pass in DDP reduces computational complexity by solving an approximated linear–quadratic regulator problem over the entire time horizon \cite{ma_car_2022}. 
Nevertheless, a major drawback of the DDP is its inability to handle hard inequality constraints. To this end, various solutions have been developed, such as active-set methods \cite{xie_active_set_2017}, quadratic barrier functions \cite{kleff_high_2021}, or augmented Lagrangian (AL) techniques \cite{Zachary_almm_2019}. Among them, the alternating direction method of multipliers (ADMM) emerged from AL garners much attention due to its decomposability. It can take advantage of the problem structure to solve large-scale optimization problems \cite{boyd_admm_2011}. The ADMM has been successfully applied in diverse robotics to reduce computation burdens, including autonomous cars \cite{ma_car_2022}, drones \cite{tinympc_2024}, and contact-aware manipulation \cite{posa_C3_2024}. However, few prior works exploit this superior efficiency to tackle robotic manipulation tasks that couple full-system dynamics with collision avoidance.

In this paper, we present a novel task-oriented MPC (ToMPC) trajectory planner based on the ADMM, which enables safe and efficient robotic manipulation in obstructed environments (see Fig. \ref{fig:framework}). The contributions are summarized as follows. 
\begin{itemize}
    \item We propose a unified MPC framework for collision-free motion and environment interaction tasks. The framework supports online planning of motion and/or force trajectories in the workspace shared with obstacles. Moreover, by exploiting full-body dynamics, compliant contact is available without relying on additional force sensors.
    \item We define a task-oriented avoidance strategy that leverages kinematic redundancy to balance obstacle avoidance and task execution. The robot link proactive motion increases the manipulation range of the end-effector. 
    \item We use ADMM distributed optimization to implement the ToMPC. The constrained nonlinear problem is separated into two subproblems solved by DDP and QP algorithms. Since equality and inequality constraints are decoupled, the MPC problem is solved in real time, and obstacle avoidance is strictly satisfied as one of hard constraints.
\end{itemize}
The approach is validated on a redundant robotic manipulator through simulations and real-hardware experiments.

The remainder of this paper is organized as follows. Section \ref{background} briefly introduces the necessary background knowledge. Section \ref{problem} details the ToMPC formulation, and Section \ref{admm} discusses its implementation through distributed optimization. Section \ref{results} shows the results validated in simulation and on real hardware. Finally, Section \ref{conclusion} summarizes this work.

\section{Background}
\label{background}
\subsection{Model Predictive Control}
    \label{back_mpc}
    MPC is an advanced control method for systems with multiple variables and constraints \cite{eth_mpc_2022}, which can be expressed as
    \begin{align}
        \label{eq:mpc_general}
        \min_{\bm{x}_{0:T}, \bm{u}_{0:T-1}}
        &\ell_T(\bm{x}_{T}) + \sum_{t=0}^{T-1} \ell_t(\bm{x}_t, \bm{u}_t) \\
        \text{s.t.} \quad 
        &\bm{x}_{t+1} = \bm{f}(\bm{x}_t, \bm{u}_t), \quad \forall t \in \{0, 1, \dots, T-1\}, \notag \\
        &\bm{c}_t(\bm{x}_t, \bm{u}_t) \geq 0, \quad \forall t \in \{0, 1, \dots, T-1\}, \notag \\
        &\bm{c}_T(\bm{x}_T) \geq 0, \quad \bm{x}_0 = \bm{x}(0), \notag
    \end{align}
    
    \noindent where $\bm{x}_t \!\in\! \mathbb{R}^{n_x}$ and $\bm{u}_t \!\in\! \mathbb{R}^{n_u}$ are the vectors of state and input variables at $t$ time step, $\bm{x}_{0:T} \!=\! (\bm{x}_0, \bm{x}_1, \cdots \!, \bm{x}_T)$ and $\bm{u}_{0:T-1} \!=\! (\bm{u}_0, \bm{u}_1, \cdots \!, \bm{u}_{T-1} )$ are the optimization variable sequences, $T$ is the prediction horizon, $\ell_t(\bm{x}_t, \bm{u}_t), \ell_T(\bm{x}_T) \!\in\! \mathbb{R}^{+} $ are the stage cost and terminal cost, $\bm{c}_t(\bm{x}_t, \bm{u}_t), \bm{c}_T(\bm{x}_T) \!\in\! \mathbb{R}^{n_c}$ are the constraint functions, and $\bm{f}(\bm{x}_t, \bm{u}_t)$ is the discretized model where the current measurement $\bm{x}(0)$ is assigned to the initial state $\bm{x}_0$. When (\ref{eq:mpc_general}) is solved iteratively, the first control input $\bm{u}_0$ from the optimal sequence is applied to the robot.

\subsection{Alternating Direction Method of Multipliers}
    \label{back_admm}
    ADMM is an AL-based method to solve optimization problems in a distributed fashion \cite{amatucci_accelerating_2024}. Consider the  following generic optimization problem:
    \begin{equation}
        \label{eq:admm_1}
        \begin{aligned}
            \min_{\bm{y}, \bm{z}} \quad
            & \ell_y(\bm{y}) + \ell_z(\bm{z}) \\
            \text{s.t.} \quad 
            & \bm{\mathcal{G}}\bm{y} - \bm{z} = 0,
        \end{aligned}
    \end{equation}
    
    \noindent where $\ell_y(\bm{y})$, $\ell_z(\bm{z})$ are the two objectives of the cost function, \ $\bm{y}$ and $\bm{z}$ are two independent decision variables associated through an equality constraint, $\bm{\mathcal{G}}$ is a projection matrix. The augmented Lagrangian function for (\ref{eq:admm_1}) is given by
    \begin{equation}
        \label{eq:admm_2}
        \begin{aligned}
        \mathcal{L}_\rho(\bm{y}, \bm{z}, \bm{\lambda}) = \ell_y(\bm{y}) \!+\! \ell_z(\bm{z}) 
        \!+\! \bm{\lambda}^\top(\bm{\mathcal{G}}\bm{y} \!-\! \bm{z})
        \!+\! \frac{\rho}{2} \| \bm{\mathcal{G}}\bm{y} \!-\! \bm{z} \|^2,
        \end{aligned}
    \end{equation}
    
    \noindent where $\bm{\lambda}$ is the Lagrangian multiplier, $\rho$ is the scalar penalty weight.  Since it is sometimes difficult to minimize both $\bm{y}$ and $\bm{z}$ in a single step, the ADMM leverages the separation of cost terms to decompose the original problem into smaller, simpler subproblems to tackle. The principle is described as
    \begin{equation}
        \label{eq:admm_3}
        \begin{aligned}
        \bm{y}^{\left[k+1\right]} &= \arg\min_{\bm{y}} \mathcal{L}_\rho (\bm{y}, \bm{z}^{\left[k\right]}, \bm{\lambda}^{\left[k\right]}), \\
        \bm{z}^{\left[k+1\right]} &= \arg\min_{\bm{z}} \mathcal{L}_\rho (\bm{y}^{\left[k+1\right]}, \bm{z}, \bm{\lambda}^{\left[k\right]}), \\
        \bm{\lambda}^{\left[k+1\right]} &= \bm{\lambda}^{\left[k\right]} + \rho(\bm{\mathcal{G}}\bm{y}^{\left[k+1\right]} - \bm{z}^{\left[k+1\right]}),
        \end{aligned}
    \end{equation}
    
    \noindent where $k$ denotes the iteration number. In (\ref{eq:admm_3}), the primal variables $\bm{y}$ and $\bm{z}$ are updated alternately, while the dual variable $\bm{\lambda}$ is updated based on their consistency error. These steps are solved in the loop until a desired convergence residual is achieved. It has been shown that the ADMM can achieve first-order convergence for convex problems \cite{boyd_admm_2011}, and exhibit favorable convergence properties for nonconvex problems, reaching acceptable solutions in fewer iterations \cite{ma_car_2022}.

\subsection{Null Space and Task Priority}
    \label{back_null}
    In $m$ dimensional task space, a redundant manipulator with $n \left(n \!>\! m \right)$ joints satisfies $\bm{\mathcal{X}} \!=\! \text{FK}(\bm{q})$, where $\bm{\mathcal{X}} \!\in\! SE(3)$ is the end-effector pose, $\bm{q} \!\in\! \mathbb{R}^n$ is the joint position, $\text{FK}(\cdot)$ is the forward kinematics. By differentiating with respect to time, the mapping between task space and joint space is given by
    \begin{equation}
        \label{eq:fw_kin}
        \bm{\mathcal{V}} = \bm{J}(\bm{q})\dot{\bm{q}},
    \end{equation}
    
    \noindent where $\bm{J} \!\in\! \mathbb{R}^{m\times n}$ is the Jacobian matrix, $\bm{\mathcal{V}} \in \mathbb{R}^m$ is the end-effector twist (or called spatial velocity) \cite{modern_robotics_2017}. The twist will contain three translational velocities and three rotational velocities if the full Cartesian space is considered ($m = 6$). Since the system is kinematically redundant, more than one configuration exists for a given end-effector pose. The additional degrees of freedom provide a null space defined by
    \begin{equation}
        \label{eq:inv_kin}
        \bm{N}(\bm{q}) = \bm{I} - \bar{\bm{J}}\bm{J}, 
    \end{equation}
    
    \noindent where $\bar{\bm{J}} = \bm{J}^\top(\bm{J} \bm{J}^\top)^{-1}$ is the pseudo-inverse of Jacobian, and $\bm{N}$ is the corresponding null space. This property allows the robot to perform multiple tasks simultaneously and prioritize them based on their importance, ensuring that low-priority tasks do not interfere with high-priority ones.

\section{Task-oriented MPC Formulation}
\label{problem}
In this section, we present the formulation of our task-oriented MPC problem in three aspects: system model, safety constraints, and cost function design, as outlined in (\ref{eq:mpc_general}).
 
\subsection{Robot and Interaction Modeling}
    The rigid body dynamics of redundant robotic manipulators can be expressed as
    \begin{equation}
        \label{eq:dyn_robot}
        \bm{M}(\bm{q})\ddot{\bm{q}} + \bm{C}(\bm{q},\dot{\bm{q}})\dot{\bm{q}} + \bm{g}(\bm{q}) = \bm{\tau} + \bm{\tau}_{ext},
    \end{equation}

    \noindent where $\bm{M} \in \mathbb{R}^{n\times n}$ the inertia matrix, $\bm{C}\in \mathbb{R}^{n\times n}$ is the Coriolis and centrifugal matrix, $\bm{g}\in \mathbb{R}^n$ is the gravity vector, $\bm{\tau} \in \mathbb{R}^n$ is the commanded joint torque, $\bm{\tau}_{ext} \in \mathbb{R}^n$ is the external torque. In the case of interaction task on the end-effector, there exists $\bm{\tau}_{ext} = \bm{J}^\top_c \bm{\mathcal{F}}$, where the external wrench or spatial force $\bm{\mathcal{F}} \in \mathbb{R}^{6}$ consists of the forces and moments that the environment applies on the end-effector, $\bm{J}_c$ is the Jacobian in the contact frame. Here, the external wrench is regarded as a state rather than a disturbance, allowing us to describe the overall system behavior. As shown in Fig. \ref{fig:principle}, an interaction model inspired by impedance control \cite{hogan_impedance_1984} is defined as
    \begin{equation}
        \label{eq:dyn_env}
        \bm{\mathcal{F}}(\bm{q}, \dot{\bm{q}}) = \bm{K}_{env}(\bm{\mathcal{X}}^{ref} \ominus \bm{\mathcal{X}}) + \bm{D}_{env}\bm{\mathcal{V}},
    \end{equation}

    \noindent where $\bm{K}_{env} \in \mathbb{R}^{6\times 6}$ and $\bm{D}_{env} \in \mathbb{R}^{6\times 6}$ are positive diagonal gain matrices representing desired stiffness and damping, the superscript $(\cdot)^{ref}$ indicates the reference, $\ominus$ denotes the difference in $SE(3)$  and converts to $\mathbb{R}^{6}$, i.e., $\bm{\mathcal{X}}_1 \ominus \bm{\mathcal{X}}_2 = \log(\bm{\mathcal{X}}^{-1}_1 \bm{\mathcal{X}}_2)$.  

    Subsequently, substituting (\ref{eq:dyn_env}) into (\ref{eq:dyn_robot}) can yield the forward dynamics (FD) with interaction ability:
    \begin{equation}
        \label{eq:forward_dyn}
        \text{FD}(\bm{q}, \dot{\bm{q}}, \bm{\tau}) = \bm{M}^{-1}( \bm{\tau} + \bm{J}^\top_c \bm{\mathcal{F}} - \bm{C}\dot{\bm{q}} - \bm{g} ) = \ddot{\bm{q}} \,.
    \end{equation}

    \noindent Let $\bm{x} \!=\! (\bm{q}, \dot{\bm{q}}) \in \mathbb{R}^{2n}$ and $\bm{u} \!=\! \bm{\tau} \in \mathbb{R}^{n}$ be the state and input variables. The state-space model in (\ref{eq:mpc_general}) is then derived as
    \begin{equation}
        \label{eq:state_space}
        \bm{x}_{t+1} = \bm{f}(\bm{x}_t, \bm{u}_t) = \bm{x}_{t} + \begin{bmatrix} \dot{\bm{q}}_t \\ \text{FD}(\bm{q}_t, \dot{\bm{q}}_t, \bm{u}_t) \end{bmatrix} \Delta t \; ,
    \end{equation}
    
    \noindent where $\Delta t$ is the time interval. (\ref{eq:state_space}) is generally nonlinear, which may cause the MPC computation to be slow or get stuck in local minima. For this reason, as stated in Section \ref{intro}, we adopt the DDP algorithm to address nonlinearity, which requires the partial derivatives of dynamics. For the interaction model (\ref{eq:dyn_env}), its derivatives with respect to the state and input are 
    \begin{equation}
        \label{eq:partial}
        \frac{\partial \bm{\mathcal{F}}}{\partial \bm{q}} = \bm{K}_{env} \bm{J}_c, \quad
        \frac{\partial \bm{\mathcal{F}}}{\partial \dot{\bm{q}}} = \bm{D}_{env} \bm{J}_c, \quad
        \frac{\partial \bm{\mathcal{F}}}{\partial \bm{u}} = 0.
    \end{equation}
    
    \noindent Other derivative terms in the dynamics can also be obtained analytically, as described in \cite{Carlos_ddp_2023}.
    
\subsection{Safety Constraints}
    \label{avoidance}
    \begin{figure}[!t]
        \centering
        \includegraphics[width=.48\textwidth]{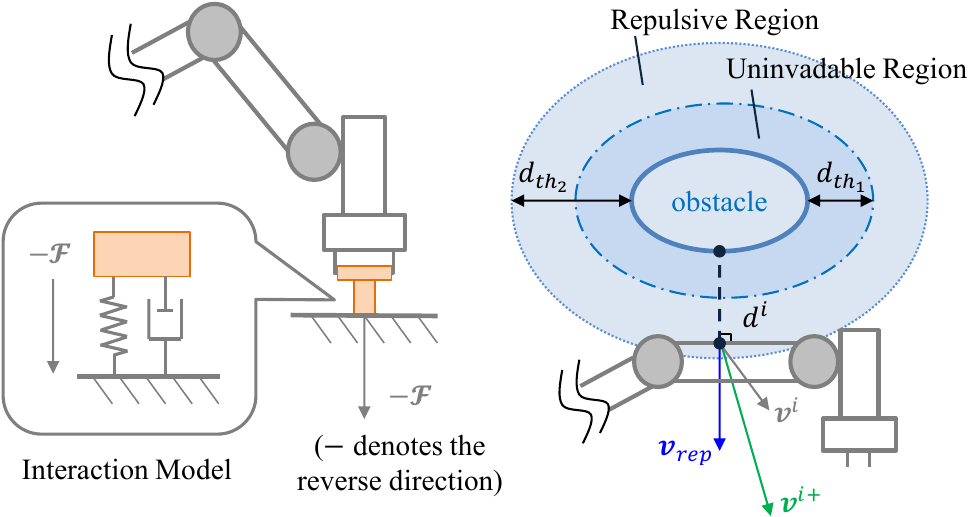}
        \caption[]{Illustration of interaction model (left) and task-oriented avoidance (right). }
        \label{fig:principle}
    \end{figure}
    To avoid collisions with obstacles in the robot workspace, the distances to obstacles are determined first. We represent all robot geometries as primitive collision bodies and apply the distance queries algorithm in \cite{Pan_fcl_2012}. The shortest distance $d^{\,i}$ of the $i$-th collision pair can be expressed as
    \begin{equation}
        \label{eq:distance}
        d^{\,i}(\bm{q}) = 
        \max_{\substack{\vec{\bm{n}}^i \in \mathbb{R}^3 \\ \|\vec{\bm{n}}^i\| = 1}} \min_{\substack{\bm{p}^i_A \in A \\ \bm{p}^i_B \in B}}  \vec{\bm{n}}^{i\top} \left( \bm{p}^i_A - \bm{p}^i_B \right),
    \end{equation}

    \noindent where $\bm{p}^i_A \in \mathbb{R}^{3}$ and $\bm{p}^i_B \in \mathbb{R}^{3}$ are the points on the robot and obstacle in the world frame, $\vec{\bm{n}}^i$ is an unit direction vector given by $\vec{\bm{n}}^i \!=\! (\bm{p}^i_A - \bm{p}^i_B) / \| \bm{p}^i_A - \bm{p}^i_B \|$. For fast computation in optimization, it is common to use linear approximation of (\ref{eq:distance}). According to \cite{Schulman_loca_2013}, the derivative of the distance function is 
    \begin{equation}
        \label{eq:dist_jacobian}
        \frac{\delta d^{\,i}}{\delta \bm{q}} \approx \pm \hat{\bm{n}}^{i\top} (\bm{J}^i_A(\bm{q}) - \bm{J}^i_B(\bm{q})), 
    \end{equation}

    \noindent where $\bm{J}^i_A$, $ \bm{J}^i_B$ are the Jacobians at the nearest points $\hat{\bm{p}}^i_A$, $\hat{\bm{p}}^i_B$, and the hat superscripts denote the optimal parameters in (\ref{eq:distance}).  Thus, the distance at the $t$th moment can be derived by
    \begin{equation}
        \label{eq:dist_eq}
         d^{\,i}(\bm{q}_t) \approx  d^{\,i}(\bm{q}) + \hat{\bm{n}}^{i\top} \bm{J}^i_A(\bm{q}) (\bm{q}_t - \bm{q}).
    \end{equation}

    \noindent Note that the computationally expensive distance query is restricted to the initial step of each MPC cycle, while subsequent states are approximated by (\ref{eq:dist_eq}). With this setup, whole-body obstacle avoidance is achieved by enforcing the following inequality constraint in the MPC:
    \begin{equation}
        \label{eq:dist_final}
        \bm{d}(\bm{q}_t) \ge \bm{d}_{th_1}, \; \forall t \in \{0, 1, \dots, T\},
    \end{equation}
    
    \noindent where $\bm{d}(\bm{q}_t) \!=\! (d^1, d^2, \cdots, d^{N_{pairs}}) \!\in\! \mathbb{R}^{N_{pairs}}$ includes all obstacle distances at $t$ time step, $N_{pairs}$ is the collision pair number, and $\bm{d}_{th_1}$ is the minimum allowable distance threshold.
    
    The hard constraint (\ref{eq:dist_final}) defines an uninvadable region around obstacles that prevents the robot from entering, see Fig.  \ref{fig:principle}. However, to avoid violating this safety constraint, the quality of manipulation may be compromised. In fact, the robot can fine-tune its movements before approaching obstacles, so that both safety and task execution are maintained simultaneously, i.e., \textbf{task-oriented avoidance}. This action is formulated as a soft constraint and will be discussed later in Section \ref{task-oriented}.

    In addition to collision avoidance, joint angle, velocity, and control input limits are also considered to ensure robot safety, given by $\bm{x}_{\min} \leq \bm{x}_t \leq \bm{x}_{\max}$ and $\bm{u}_{\min} \leq \bm{u}_t \leq \bm{u}_{\max}$.

\subsection{Cost Function Design}
    The objective of the ToMPC is to generate feasible motion trajectories in obstructed environments, and/or plan compliant interaction force for contact scenarios. Besides, it is expected to enhance manipulation efficiency by minimizing the impact of obstacle avoidance on task execution. To this end, the stage and terminal cost functions are designed as follows:
    \begin{align}
        \label{eq:cost_function}
        &\ell_t(\bm{x}_t, \bm{u}_t) = \ell_m(\bm{x}_t) \!+\! \ell_f(\bm{x}_t) \!+\! \ell_{rep}(\bm{x}_t) \!+\! \ell_u(\bm{x}_t, \bm{u}_t), \\
        &\ell_T(\bm{x}_T) = \ell_m(\bm{x}_T) + \ell_f(\bm{x}_T), 
    \end{align}

    \noindent where $\ell_m(\bm{x}_t)$ and $\ell_f(\bm{x}_t)$ correspond to the cost terms for motion control and force control tasks, $\ell_{rep}(\bm{x}_t)$ is for task-oriented avoidance, $\ell_u(\bm{x}_t, \bm{u}_t)$ penalizes the control input. 
    
    \subsubsection{Motion and Force Tracking}
        We define the following cost function to track motion trajectories in Cartesian space:
        \begin{equation}
            \label{eq:motion_cost}
            \ell_{m}(\bm{x}_t) = \| \bm{\mathcal{X}}^{ref} \ominus \bm{\mathcal{X}}(\bm{q}_t) \|^2_{\bm{Q}_{m}} + \| \dot{\bm{q}}_t \|^2_{\bm{Q}_{s}},
        \end{equation}
        
        \noindent where the first term regulates the end-effector to the reference pose, while the second term penalizes the joint velocity to ensure trajectory smoothness and stabilize joint motion in null space. $\bm{Q}_{m} \!\in\! \mathbb{R}^{6\times 6}$ and $\bm{Q}_{s} \!\in\! \mathbb{R}^{n\times n}$ are the positive diagonal weighting matrices. Similarly, the cost function of  interaction force is
        \begin{equation}
            \label{eq:force_cost}
            \ell_{f}(\bm{x}_t) = \| \bm{\mathcal{F}}^{ref} -  \bm{\mathcal{F}}(\bm{q}_t, \dot{\bm{q}}_t) \|^2_{\bm{Q}_{f}} ,
        \end{equation}

        \noindent where the wrench $\bm{\mathcal{F}}$ is computed by (\ref{eq:dyn_env}), and $\bm{Q}_{f} \!\in\! \mathbb{R}^{6\times 6}$ is the weighting matrix. Notably, the elements of $\bm{Q}_{m}$, $\bm{Q}_{f}$ must be chosen in a complementary manner, such that each Cartesian direction is either motion- or force-controlled.
        
    \subsubsection{Task-oriented Avoidance}
        \label{task-oriented}
        For this avoidance behavior, we assume a repulsive region with a distance threshold $d_{th_2} (d_{th_2} > d_{th_1})$ around obstacles, as illustrated in Fig. \ref{fig:principle}. When the link $i$ enters into this region, a virtual repulsive velocity $\bm{v}_{rep} \in \mathbb{R}^3$ is generated along $\hat{\bm{n}}^i$ at the closest point:
        \begin{equation}
            \label{eq:v_rep}
            \bm{v}_{rep} = k_{rep} (d_{{th}_2} - d^{\,i}) \hat{\bm{n}}^i ,
        \end{equation}
    
        \noindent where $k_{rep}$ is a positive scaling factor. It can be seen that as the link penetrates deeper, the magnitude of $\bm{v}_{rep}$ increases and reaches the maximum at the boundary of uninvadable region. Given the current velocity $\bm{v}^i$, the resulting total velocity $\bm{v}^{i+}$ is
        \begin{equation}
            \label{eq:v_total}
            \bm{v}^{i+} = \bm{v}^i + \bm{v}_{rep}.
        \end{equation}
    
        \noindent With this repulsive motion, the robot can actively maneuver its links away from obstacles. However, this behavior may interfere with the  execution of main tasks described in (\ref{eq:motion_cost}) or (\ref{eq:force_cost}). To harmonize the conflicting objectives, we leverage kinematic redundancy, incorporating task-oriented avoidance into the cost function as a soft constraint:
        \begin{equation}
            \label{eq:cost_rep}
            \ell_{rep}(\bm{x}_t) \!=\! 
            \begin{cases}
                \| \bm{N}_{task} (\dot{\bm{q}}_t \!-\! \bar{\bm{J}}^i_A \bm{v}^{i+}) \|^2_{\bm{Q}_{rep}}, \!\!\!\!\!& \text{if } d^{\,i} \leq d_{{th}_2} \\
                0, & \text{otherwise }
            \end{cases},
        \end{equation}

        \noindent where $\bm{Q}_{rep} \!\in\! \mathbb{R}^{n \times n}$ is the weighting matrix, and $\bm{N}_{task} \!=\! \bm{I} - \bar{\bm{J}}_{task}\bm{J}_{task}$ represents the null space of the task Jacobian $\bm{J}_{task} \in \mathbb{R}^{m \times n}$ as defined in (\ref{eq:inv_kin}). 
        
        Additionally, we introduce a goal relaxation term $\xi \in \left[0, 1\right]$ to modulate the end-effector manipulation:
        \begin{equation}
            \label{eq:lambda}
            \xi(d^{\,i}) =
            \begin{cases}
                \exp( - \frac{\alpha(d_{{th}_2} - d^{\,i})} {d_{{th}_2} -  d_{{th}_1}}), & \text{if } d^{\,i} \leq d_{{th}_2} \\
                1, & \text{otherwise }
            \end{cases},
        \end{equation}
    
        \noindent where $\alpha$ is a positive shaping factor. Using this function, the goal weighting matrices are updated as $\bm{Q}_{m} \!\leftarrow\! \xi(d^{\,i}) \bm{S} \bm{Q}_{m}$ and $\bm{Q}_{f} \leftarrow \xi(d^{\,i}) \bm{S} \bm{Q}_{f}$, where $\bm{S}$ is a selection matrix that isolates all elements irrelevant to manipulation. This revision in motion and force tracking costs will decrease the contribution of goal attraction as the robot is close to obstacles. 

    \subsubsection{Control Regularization}
        We regulate the control input to the torque that compensates for gravity at a given configuration: 
        \begin{equation}
            \label{eq:control_cost}
            \ell_{m}(\bm{x}_t, \bm{u}_t) = \| \bm{u}_t - \bm{g}(\bm{q}_t) \|^2_{\bm{R}},
        \end{equation}
        
        \noindent where $\bm{R} \in \mathbb{R}^{n \times n}$ is the control input weighting matrix.

\section{Distributed Optimization}
\label{admm}
The ToMPC planner unifies the tasks of collision-free motion and environment interaction, providing a holistic framework to robot control. However, this integration leads to a complex constrained nonlinear optimization problem. In this section, we use the ADMM method to efficiently implement the problem, especially to solve the coupling in kinematics and dynamics.

According to Section \ref{problem}, the ToMPC is summarized as 
\begin{subequations}
    \label{eq:mpc_all}
    \begin{align}
        \min_{ \substack{ \bm{x}_{0:T} \\ \bm{u}_{0:T-1} } } \quad
        &\ell_T(\bm{x}_{T}) + \sum_{t=0}^{T-1} \ell_t(\bm{x}_t, \bm{u}_t) \\
        \text{s.t.} \quad 
        &\bm{x}_{t+1} = \bm{f}(\bm{x}_t, \bm{u}_t), \quad \forall t \in \{0, 1, \dots, T-1\}, \label{eq:mpc_f} \\
        &\bm{u}_{\min} \leq \bm{u}_t \leq \bm{u}_{\max}, \; \forall t \in \{0, 1, \dots, T-1\}, \label{eq:mpc_u} \\
        &\bm{x}_{\min} \leq \bm{x}_t \leq \bm{x}_{\max}, \; \forall t \in \{0, 1, \dots, T\}, \label{eq:mpc_x} \\
        &\bm{d}_{th_1} \leq \bm{d}(\bm{q}_t), \; \forall t \in \{0, 1, \dots, T\}, \label{eq:mpc_dist} 
    \end{align}
\end{subequations}

\noindent where the first state $\bm{x}_0$ is updated by the joint measurement before each call to MPC. For brevity, we define the indicator function with respect to a set $\mathcal{C}$ as
\begin{equation}
    \label{eq:indicator}
    \mathcal{I}_\mathcal{C}(c) =
    \begin{cases}
        0, & \text{if } c \in \mathcal{C} \\
        +\infty, & \text{otherwise }
    \end{cases}.
\end{equation}

\noindent Thus, the optimization problem with multiple constraints in (\ref{eq:mpc_all}) can be equivalently expressed as follows:
\begin{align}
    \label{eq:mpc_indicator}
    \min_{\bm{y}, \bm{z}} \quad
    &\ell_T(\bm{x}_{T}) + \sum_{t=0}^{T-1} \ell_t(\bm{x}_t, \bm{u}_t) + \mathcal{I}_{\mathcal{Y}}(\bm{y}) + \mathcal{I}_{\mathcal{Z}}(\bm{z}) \notag \\
    \text{s.t.} \quad 
    &\bm{\mathcal{G}}\bm{y} - \bm{z} = 0,
\end{align}

\noindent where $\bm{y} = \left( \bm{u}_0, \bm{x}_1, \dots, \bm{u}_{T-1}, \bm{x}_T \right) \in \mathbb{R}^{(n_x + n_u)T}$ denotes the sequence of decision variables, subject to the equality constraint set $\mathcal{Y}$, and $\bm{z}$ is obtained by applying the projection matrix $\bm{\mathcal{G}}$ to $\bm{y}$, subject to the inequality constraint set $\mathcal{Z}$. Referring to (\ref{eq:admm_2}), the augmented Lagrangian function is then derived as
\begin{align}
    \label{eq:admm_indicator}
    \mathcal{L}_\rho(\bm{y}, \bm{z}, \bm{\lambda}) 
    &= \ell_T(\bm{x}_{T}) + \sum_{t=0}^{T-1} \ell_t(\bm{x}_t, \bm{u}_t) + \mathcal{I}_{\mathcal{Y}}(\bm{y}) + \mathcal{I}_{\mathcal{Z}}(\bm{z}) \notag \\
    &\quad + \frac{\rho}{2} \| \bm{\mathcal{G}}\bm{y} - \bm{z} + \rho^{-1}\bm{\lambda}\|^2 
    - \frac{1}{2\rho}\| \bm{\lambda} \|^2.
\end{align}

\noindent This form of problem can be solved using the ADMM algorithm, as described in (\ref{eq:admm_3}). In the following, we provide a detailed formulation of the first and second steps of the algorithm.

\subsection{Task Execution}
    The first ADMM step focuses on the execution of manipulation tasks. $\bm{y}$ is set as the optimization variable and updated after solving, while $\bm{z}$ and $\bm{\lambda}$ are fixed and treated as constants:
    \begin{align}
        \label{eq:admm_step_1}
        \bm{y}^{\left[k+1\right]}
        &= \arg\min_{\bm{y}} \;
        \ell_T(\bm{x}_{T}) + \sum_{t=0}^{T-1} \ell_t(\bm{x}_t, \bm{u}_t) + \mathcal{I}_{\mathcal{Y}}(\bm{y}) \notag \\
        &\quad + \frac{\rho}{2} \| \bm{\mathcal{G}}\bm{y} - \bm{z}^{\left[k\right]} + \rho^{-1}\bm{\lambda}^{\left[k\right]}\|^2.
    \end{align}

    \noindent For the convenience of deployment, (\ref{eq:admm_step_1}) is rewritten as
    \begin{align}
        \label{eq:admm_step_1_spec}
        \min_{ \bm{y} } \quad
        &\ell_T(\bm{x}_T) + \sum_{t=0}^{T-1} \Bigg( \ell_m(\bm{x}_t) + \ell_f(\bm{x}_t) + \ell_u(\bm{x}_t, \bm{u}_t) \Bigg. \notag \\
        &\Bigg.+ \frac{\rho}{2} 
         \left \| \bm{\mathcal{G}}_t \bm{y}_t - \bm{z}^{\left[k\right]}_t
         + \rho^{-1} \bm{\lambda}^{\left[k\right]}_t \right \|^2 \Bigg) 
        \quad \text{s.t.} \;
         (\ref{eq:mpc_f}),
    \end{align}
    \noindent where the cost function includes terms related to the motion task $\ell_m(\bm{x}_t)$, force task $\ell_f(\bm{x}_t)$, and control regularization $\ell_u(\bm{x}_t, \bm{u}_t)$. The subscript $(\cdot)_t$ in $\bm{\mathcal{G}}$, $\bm{y}$, $\bm{z}$, and $\bm{\lambda}$ indicates the value of each variable at time step $t$. Since only equality constraints are incorporated, this nonlinear subproblem can be efficiently solved by the DDP algorithm.
    
\subsection{Safety Guarantee}
    In the second ADMM step, robot safety during manipulation is guaranteed. Given constant $\bm{y}^{\left[k+1\right]}$ and $\bm{\lambda}^{\left[k\right]}$, the subproblem involving the optimization variable $\bm{z}$ becomes
    \begin{align}
        \label{eq:admm_step_2}
        \bm{z}^{\left[k+1\right]}
        &= \arg\min_{\bm{z}} \;
        \ell_T(\bm{x}_{T}) + \sum_{t=0}^{T-1} \ell_t(\bm{x}_t, \bm{u}_t) + \mathcal{I}_{\mathcal{Z}}(\bm{z}) \notag \\
        &\quad + \frac{\rho}{2} \| \bm{\mathcal{G}}\bm{y}^{\left[k+1\right]} - \bm{z} + \rho^{-1}\bm{\lambda}^{\left[k\right]}\|^2.
    \end{align}
    
    \noindent We note that (\ref{eq:admm_step_2}) can be split into $T$ smaller subproblems that are solved in parallel; i.e., for each $t$ time step, it has
    \begin{align}
        \label{eq:admm_step_2_spec}
        \min_{ \bm{z}_t } \quad
        &\ell_{rep}(\bm{x}_t) + \frac{\rho}{2} 
         \left \| \bm{\mathcal{G}}_t\bm{y}^{\left[k+1\right]}_t - \bm{z}_t
         + \rho^{-1} \bm{\lambda}^{\left[k\right]}_t   \right \|^2 \notag \\
        \text{s.t.} \quad 
        & (\ref{eq:mpc_u})-(\ref{eq:mpc_dist}),
    \end{align}
    \noindent where $\ell_{rep}(\bm{x}_t)$ is the cost term of task-oriented avoidance. To further simplify, the constraint $(\ref{eq:mpc_u})$ can be extracted in closed form, and the rest becomes a convex QP problem with respect to the state, which can be quickly solved using a general QP solver.

\subsection{Algorithm Overview}

    The ToMPC developed on the ADMM method is summarized in Algorithm \ref{alg:admm_tompc}. For fast convergence at each MPC cycle, the solution from last cycle is used as a warm start of the current cycle, i.e., $(\bm{x}^{\left[0\right]}_{0:T}, \bm{u}^{\left[0\right]}_{0:T-1}, \bm{\lambda}^{\left[0\right]}) \!\leftarrow\! (\bm{x}^{\left[k\right]}_{0:T}, \bm{u}^{\left[k\right]}_{0:T-1}, \bm{\lambda}^{\left[k\right]})$. Additionally, we define the ADMM stopping criterion as 
    \begin{equation}
        \label{eq:admm_step_3}
        r^{\left[k\right]} = \| \bm{\mathcal{G}}\bm{y}^{\left[k\right]} - \bm{z}^{\left[k\right]} \|_\infty \leq r_{th} , 
    \end{equation}
    \noindent where $r^{\left[k\right]}$ denotes the infinity norm of the residual, and $r_{th}$ is a positive scalar threshold. It is worth highlighting that the distributed structure divides the cost terms as well as the constraints. The dynamic constraints are implicitly satisfied by the DDP, while the inequality constraints are enforced through the QP, leveraging the numerical strengths of different solvers. This separation allows the ToMPC to plan position, velocity, and torque commands in real time.  In practice, a proportional-derivative (PD) controller is designed to pack these commands:
    \begin{equation}
        \label{eq:PD}
        \bm{\tau}^{des} = \bm{\tau}_{mpc} + \bm{K}_{p}(\bm{q}^{des} - \bm{q}) + \bm{K}_{d}(\dot{\bm{q}}^{des} - \dot{\bm{q}}),
    \end{equation}
    
    \noindent where $\bm{\tau}^{des}$ is the final torque command, $\bm{\tau}_{mpc}$, $\bm{q}^{des}$, $\dot{\bm{q}}^{des}$ are calculated by the ToMPC, $\bm{K}_{p}$, $\bm{K}_{d}$ are the positive diagonal gain matrices. In (\ref{eq:PD}), the PD term is able to compensate for some model errors such as joint friction, but we need to emphasize that it contributes little to the total torque. The actuator behavior is still mainly determined by the feedforward term $\bm{\tau}_{mpc}$.
    
    \begin{algorithm}[t]
        \caption{ADMM-based ToMPC}
        \label{alg:admm_tompc}
        \begin{algorithmic}[1]
        \STATE \textbf{Initialize:} $
        \bm{x}^{\left[0\right]}_{0:T}, \; 
        \bm{u}^{\left[0\right]}_{0:T-1}, \; \bm{\lambda}^{\left[0\right]}, \; 
        \bm{x}^{\left[0\right]}_0 \!\leftarrow\! \bm{x}(0), \; 
        k \!\leftarrow\! 0
        $
        \STATE $\bm{y}^{\left[0\right]}, \bm{z}^{\left[0\right]} \leftarrow \bm{x}^{\left[0\right]}_{1:T}, \; \bm{u}^{\left[0\right]}_{0:T-1}$
        \REPEAT
            \STATE $\bm{y}^{\left[k+1\right]} = \arg\min_{\bm{y}} \mathcal{L}_\rho( \bm{y}, \; \bm{z}^{\left[k\right]}, \; \bm{\lambda}^{\left[k\right]} )$ \hfill \# DDP (\ref{eq:admm_step_1_spec})
            \FOR{$t = 0, 1, \cdots, T\!-\!1$}
                \STATE $\bm{z}^{\left[k+1\right]}_t = \arg\min_{\bm{z}_t} \mathcal{L}_\rho( \bm{y}^{\left[k+1\right]}_t, \; \bm{z}_t, \;
                \bm{\lambda}^{\left[k\right]}_t )$ \hfill \# QP (\ref{eq:admm_step_2_spec})
            \ENDFOR
            \STATE $\bm{\lambda}^{\left[k+1\right]} = \bm{\lambda}^{\left[k\right]} + \rho(\bm{\mathcal{G}}\bm{y}^{\left[k+1\right]} - \bm{z}^{\left[k+1\right]})$
            \STATE $k \leftarrow k+1$
        \UNTIL{stopping criterion is satisfied}
        \STATE $\bm{q}^{des}, \; \dot{\bm{q}}^{des}, \; \bm{\tau}_{mpc} \leftarrow \bm{x}^{\left[k\right]}_1, \; \bm{u}^{\left[k\right]}_0$
        \RETURN $\bm{q}^{des}, \; \dot{\bm{q}}^{des}, \; \bm{\tau}_{mpc}$
        \end{algorithmic} 
    \end{algorithm}
    
\begin{figure}[t]
    \vspace{-0.4cm}
    \centering
    \subfloat{\includegraphics[width=.16\textwidth]{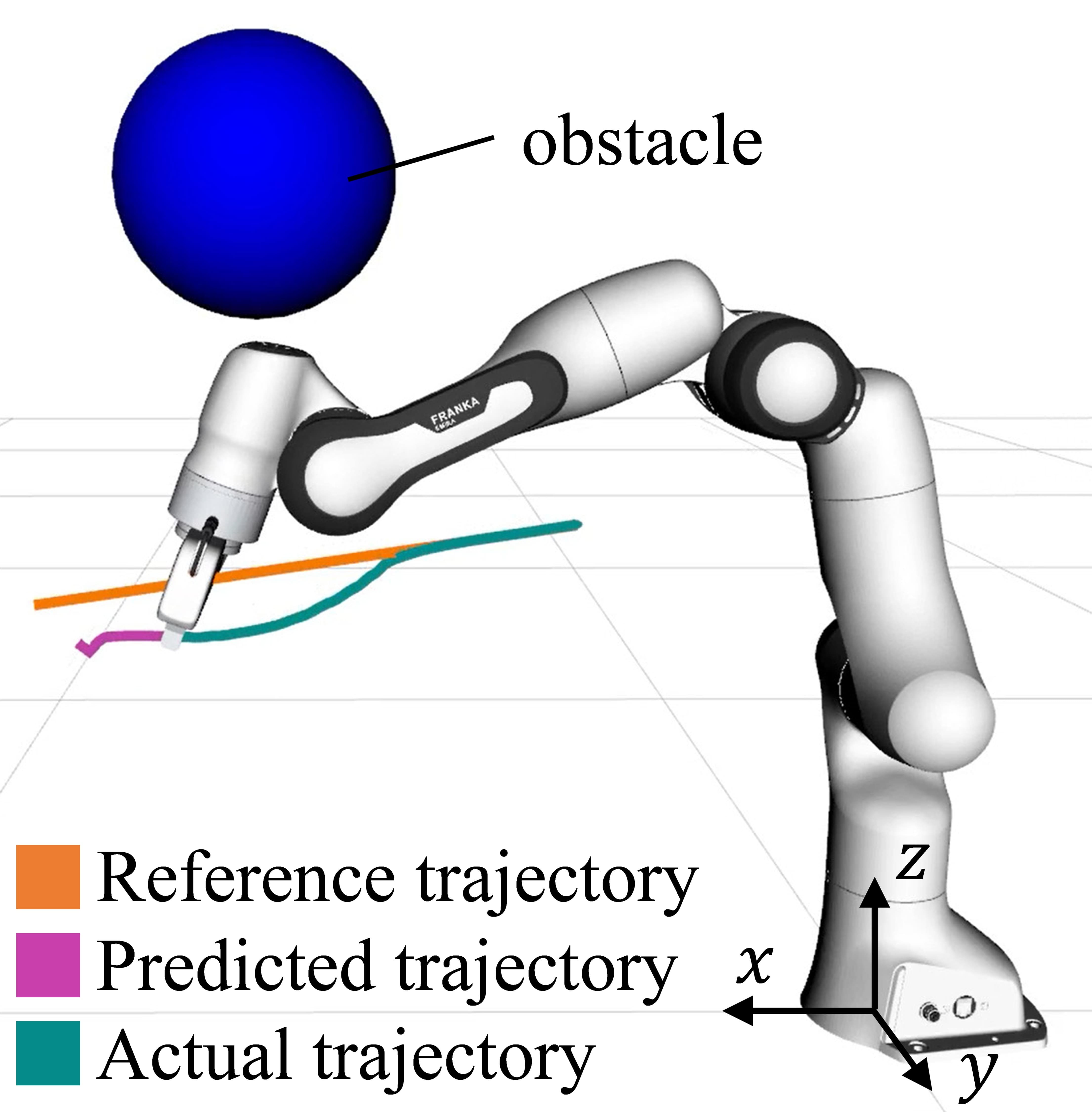}}
    \hfill
    \subfloat{\includegraphics[width=.16\textwidth]{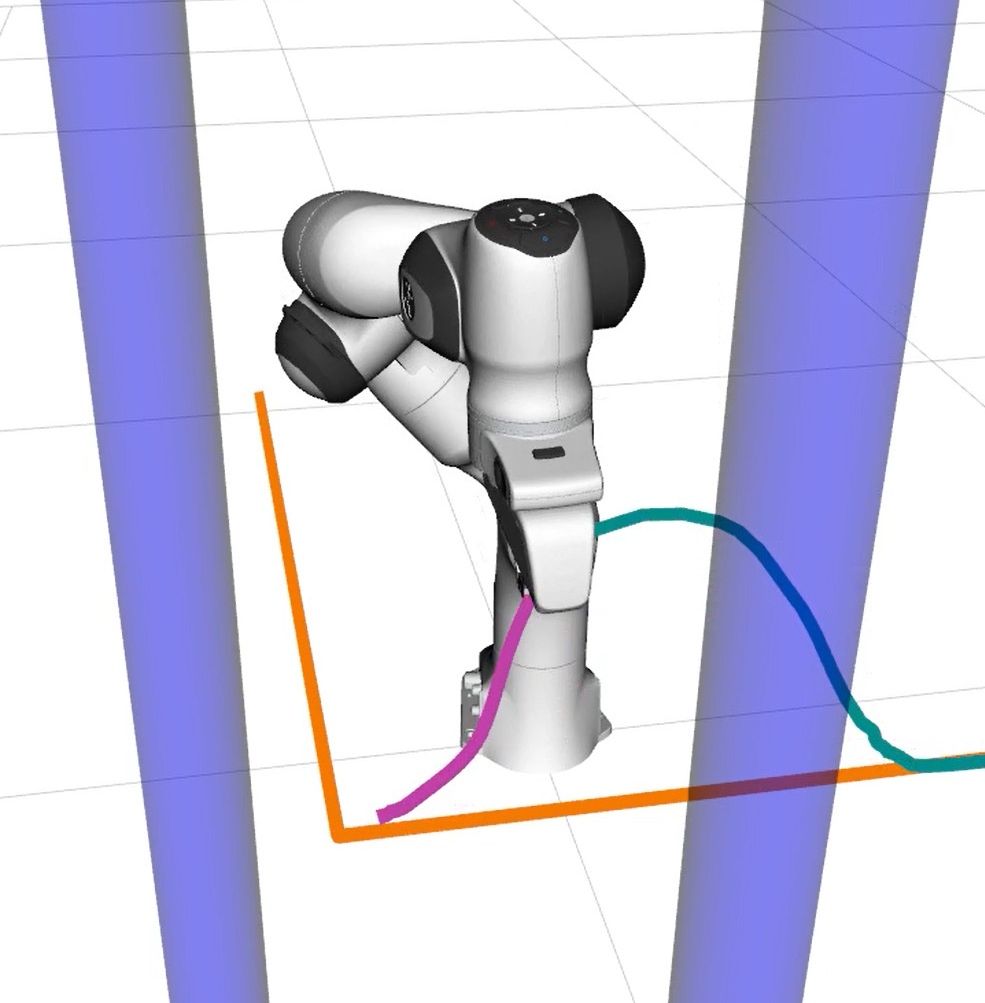}}
    \hfill
    \subfloat{\includegraphics[width=.16\textwidth]{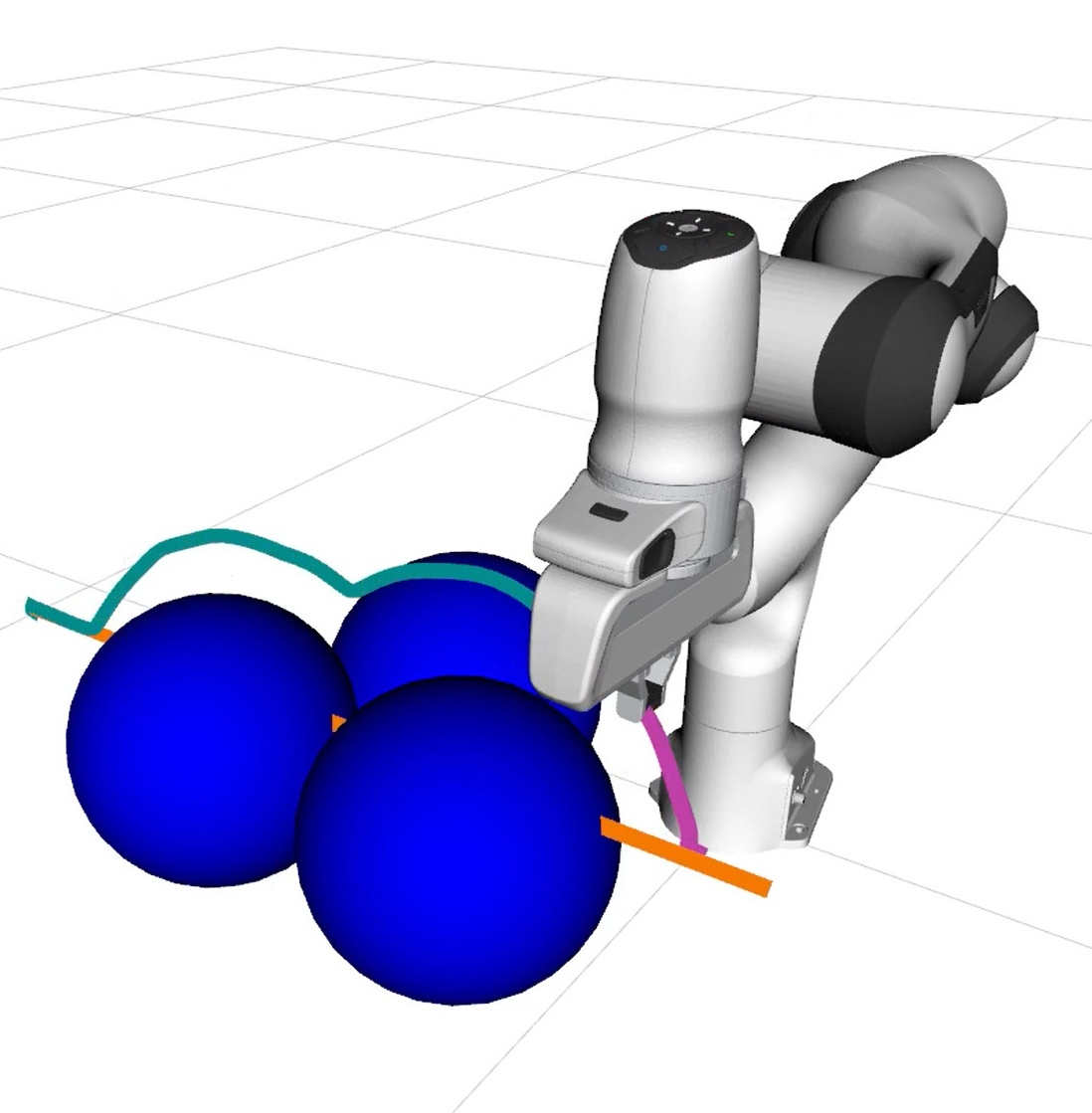}} \\
    \caption[Simulation]{Scenario $\#$1: Simulation of the robot performing obstacle avoidance with various shapes and quantities of surrounding objects. }
    \label{fig:1_sim}
    \vspace{-0.2cm}
\end{figure}

\section{Simulations and Experiments}
\label{results}
In this section, we validate the proposed ToMPC algorithm through simulations and experiments on a Franka Emika Panda robot. The robot has seven torque-controllable joints and can be equipped with a two-finger gripper for manipulation tasks. The Pinocchio library \cite{pinocchio_2019} is utilized for efficient computation of robot dynamics and obstacle distances. The FDDP \cite{Carlos_ddp_2023} and qpOASES \cite{qpoases_2014} solvers are employed to achieve the two ADMM steps in our ToMPC planner. All algorithms are implemented in C++17 and executed on a computer with an Intel Core i9-9900K CPU, running Ubuntu 20.04 with ROS Noetic. The generated torque commands are applied at a control frequency of 1 kHz.

\subsection{Scenario \#1: Obstacle Avoidance}
    \begin{figure}[t]
        \centering
        \subfloat{\includegraphics[width=.48\textwidth]{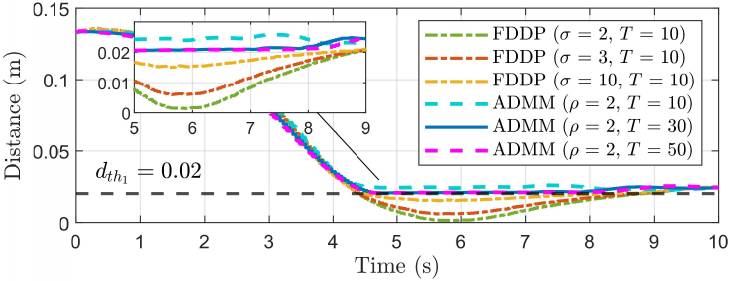}} \\
        \subfloat{\includegraphics[width=.48\textwidth]{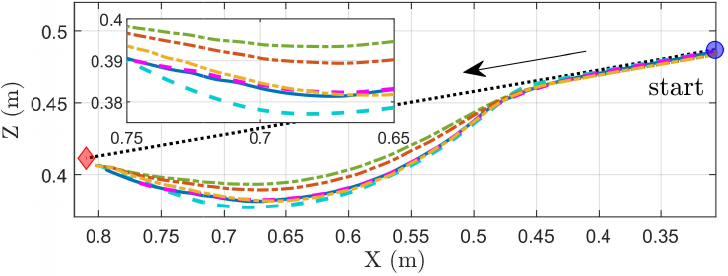}} \\
        \caption[Scenario-1]{Simulation results for Scenario \#1. \textit{Top:} Minimum distances to the obstacle (Fig. \ref{fig:1_sim} \textit{Left}). The FDDP and ADMM methods are compared under different parameter settings. \textit{Bottom:} Cartesian trajectories for all six test cases. }
        \label{fig:1_sim_results}
        \vspace{-0.2cm}
    \end{figure}

    \begin{table}[!t]
        \centering
        \caption{Minimum Distance and Computation Time in Scenario \#1} 
        \vspace{-0.2cm}
        \label{table:fddp_vs_admm}
        \begin{threeparttable}
            \setlength{\tabcolsep}{5pt} 
            \renewcommand{\arraystretch}{1.2} 
            \begin{tabular}{c c c c c c c}
                \toprule
                \textbf{Method} & $\sigma$ or $\rho$ & $T$ & \textbf{Min Dist} (m) & \textbf{Max Time} (s) & \textbf{Avg Time} (s) \\
                \hline
                \multirow{3}*{ \shortstack{FDDP \\ \cite{Carlos_ddp_2023}} }
                        & $2$ & $10$ & $0.0014$ & $0.0275$ & $0.0091$ \\
                        & $3$ & $10$ & $0.0060$ & $0.0246$ & $0.0093$ \\
                        & $10$& $10$ & $0.0151$ & $0.0288$ & $0.0108$ \\
                \hline
                \multirow{3}*{ \shortstack{ADMM \\ (ours)} }
                        & $2$ &  $10$ & $0.0229$ & $0.0013$ &$0.0009$ \\
                        & $\bf{2}$ & $\bf{30}$  & $\bf{0.0205}$ & $\bf{0.0037}$ & $\bf{0.0025}$\\
                        & $2$ &  $50$ & $0.0205$ & $0.0059$ & $0.0043$ \\
                \bottomrule
            \end{tabular}
        \end{threeparttable}
        \vspace{-0.2cm} 
    \end{table}
    
    The first scenario is performed in simulation to demonstrate the numerical behavior of our ADMM solver. Fig. \ref{fig:1_sim} illustrates the robot leveraging the proposed method to avoid obstacles of various shapes and quantities. A case involving dynamic obstacles is also considered, as shown in the accompanying video. The task-oriented avoidance is deactivated here to simplify the subsequent comparison. To highlight the advantage of ADMM, the original FDDP solver is applied to the same MPC problem including a spherical obstacle (Fig. \ref{fig:1_sim} \textit{Left}). Since the FDDP cannot handle hard constraints on state variables, obstacle avoidance is achieved by incorporating quadratic barrier terms into the cost function, which is defined as follows:
    \begin{equation}
        \label{eq:fddp}
        \ell_{avd}(\bm{x}_t) =
        \begin{cases}
            \frac{\sigma}{2} \sum_{i=0}^{N_{pairs}} ( d^i(\bm{q}_t) \!-\! d_{{th}_1}  )^2, & \text{if } d^i(\bm{q}_t) \!\leq\! d_{{th}_1} \\ 
            0, & \text{otherwise} \notag \\
        \end{cases},
    \end{equation}

    \noindent where $\sigma \in \mathbb{R}^{+}$ is the cost weight. Fig. \ref{fig:1_sim_results} plots the comparative results under different parameter settings. It is clearly seen that the minimum obstacle distances for all FDDP cases fall below the threshold $d_{th_1}=0.02$ m, indicating a violation of the safety constraint. A higher weight $\sigma$ penalizes the collision cost more heavily, but it may cause system instabilities by interfering with the task objectives. In contrast, with identical weights on the task cost terms, the ADMM method can strictly satisfy the distance limitation. This superior performance is attributed to the use of hard constraints and Lagrange multipliers, the latter of which iteratively reduce constraint violations and guide the solution toward feasibility. 
    In addition, we notice that the proposed method requires less computation time even with a longer horizon $T$, see Table \ref{table:fddp_vs_admm}. This advantage arises from the ADMM-based separation, which extracts inequality constraints from the DDP solver and employs the linear approximation in (\ref{eq:dist_eq}) to simplify obstacle avoidance for applying in the QP solver. We select $T = 30$ for the following scenarios, corresponding to a planning horizon $1.5$ s with a step size $\Delta t = 0.05$ s.

\subsection{Scenario \#2: Task-oriented Avoidance}
    \begin{figure}[!t]
        \vspace{-0.3cm}
        \centering
        \subfloat{\includegraphics[width=.24\textwidth]{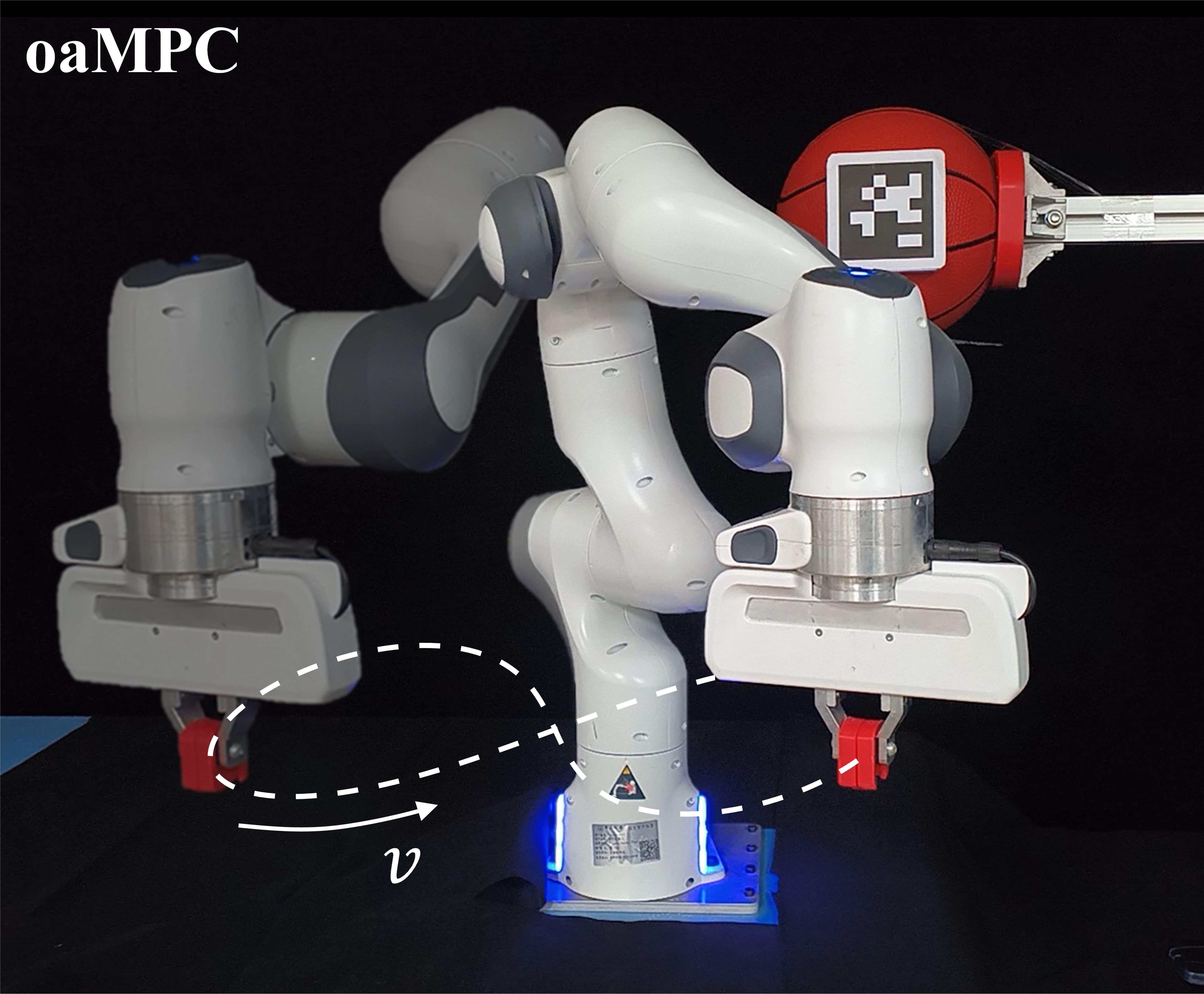}}
        \hfill
        \subfloat{\includegraphics[width=.24\textwidth]{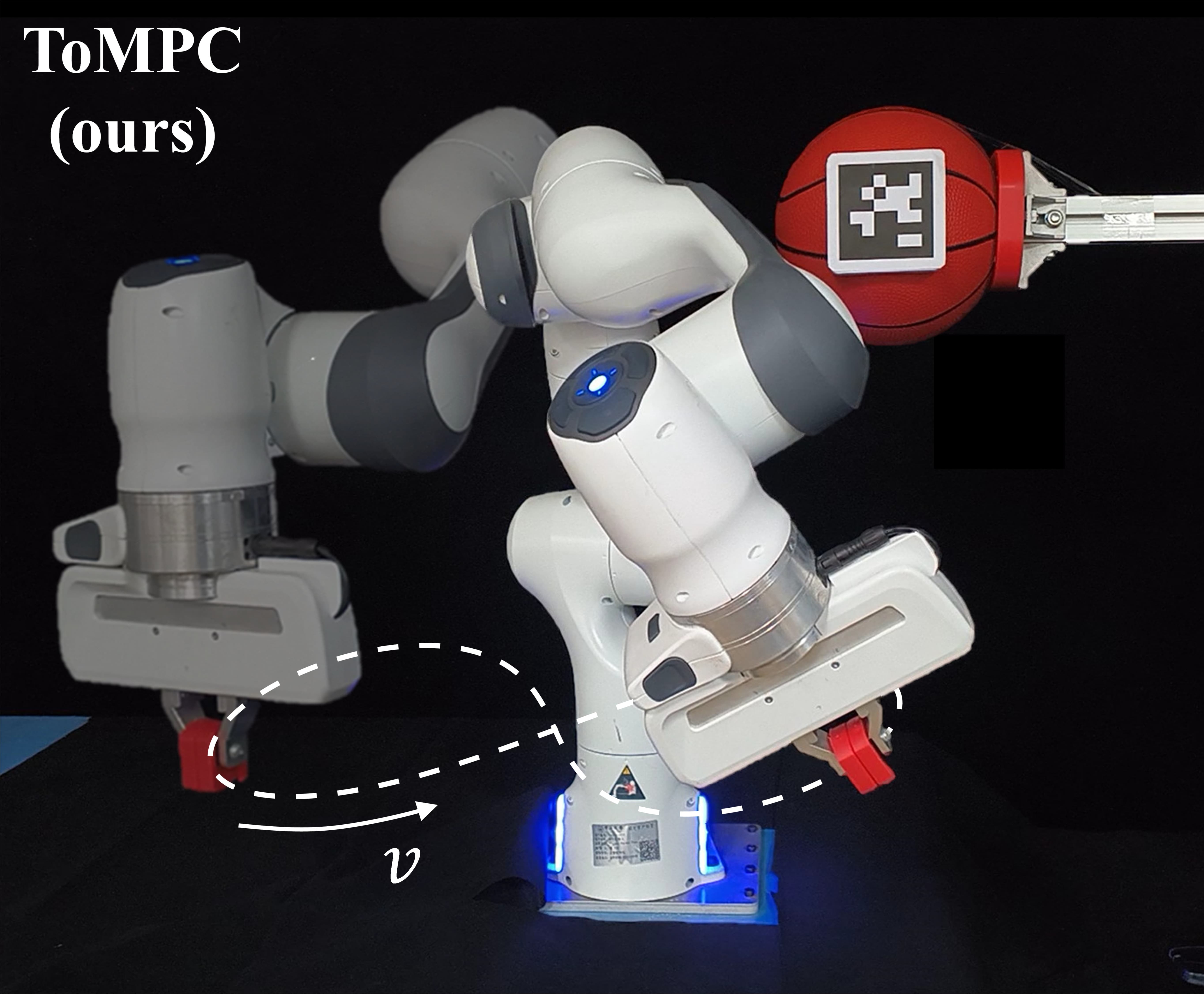}} \\
        \caption[Scenario-2]{Scenario \#2: The robot manipulates a red block while avoiding a ball obstacle. The two diagrams are experimental snapshots of conventional avoidance (oaMPC) and task-oriented avoidance (ToMPC), respectively. }
        \label{fig:2_exp}
        \vspace{-0.3cm}
    \end{figure}
    
    \begin{figure}[!t]
        \centering
        \subfloat{\includegraphics[width=.48\textwidth]{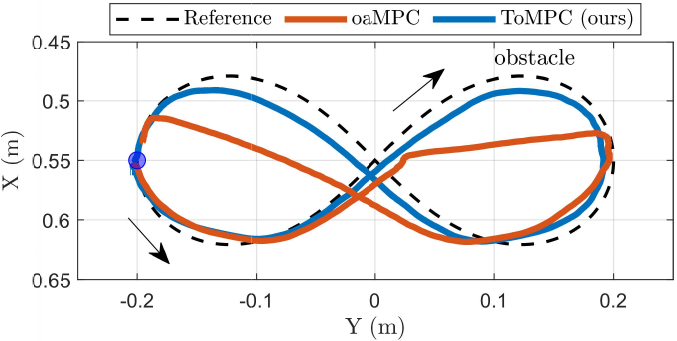}} \\
        \vspace{-0.3cm}
        \subfloat{\includegraphics[width=.48\textwidth]{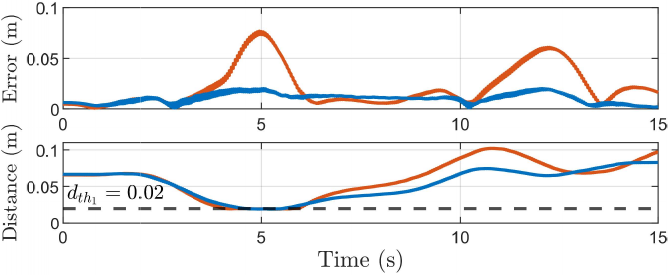}} \\
        \vspace{-0.3cm}
        \caption[Scenario-2-result]{Experimental results for Scenario \#2. \textit{Top:} Cartesian trajectory in the X-Y plane. \textit{Middle:} Norm of Cartesian trajectory errors. \textit{Bottom:} Minimum distance to the ball obstacle. }
        \label{fig:2_exp_results}
        \vspace{-0.3cm}
    \end{figure}

\begin{figure}[!t]
        \vspace{-0.3cm}
        \centering
        \subfloat{\includegraphics[width=.48\textwidth]{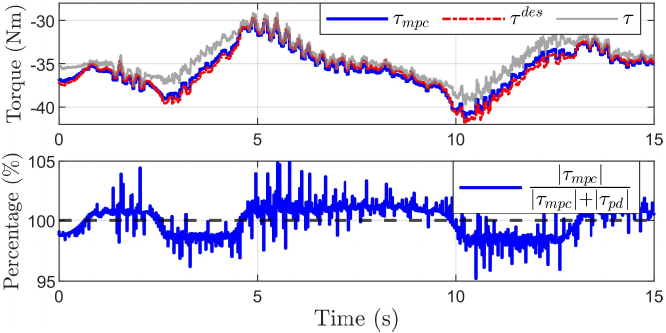}} \\
        \caption[Sigmoid]{Joint torque of the ToMPC in Scenario \#2. \textit{Top:} MPC torque $\tau_{mpc}$, final desired torque $\tau^{des}$, and measured torque $\tau$ for the second joint. \textit{Bottom:} Percentage of the feedforward torque in the total command. }
        \label{fig:2_torque}
        \vspace{-0.3cm}
    \end{figure}

    In the second scenario, the capability of task-oriented avoidance in obstructed environments is experimentally validated on hardware. As presented in Fig. \ref{fig:2_exp}, the robot gripper manipulates a red block along a Lemniscate of Bernoulli in Cartesian space, while a ball marked with a tag represents the workspace obstacle. The ball's position relative to the robot base is predetermined using the ViSP package \cite{ViSP_2005}. In this experiment, the conventional avoidance strategy similar to \cite{eth_mpc_2022} is used as the benchmark. The underlying MPC follows the same architecture as the proposed ToMPC (Algorithm \ref{alg:admm_tompc}), but with the task-oriented mechanism disabled by setting the cost term $\ell_{rep}(\bm{x}_t)$ to zero and fixing the goal relaxation factor $\xi(d^i)$ at one. For clarity, we refer to this obstacle-avoidance MPC as \textbf{oaMPC}. 
    The experimental snapshots reveal that when the robot links approach the ball, the proposed ToMPC allows the robot to twist its whole body to optimize the goal-reaching task, whereas the benchmark exhibits larger Cartesian tracking errors despite ensuring robot safety.
    Fig. \ref{fig:2_exp_results} records the actual Cartesian trajectories in the X-Y plane, along with the motion error norm and obstacle distance. Compared to the benchmark, the ToMPC preserves task execution even though strictly enforcing a minimum distance above the threshold $d_{th_1}$ during the approach phase (4-6 s).
    Furthermore, Fig. \ref{fig:2_torque} depicts the joint torque profiles under the ToMPC, where the MPC torque $\tau_{mpc}$, final desired torque $\tau^{des}$, and measured torque $\tau$ for the second joint are compared. The percentage $|\tau_{mpc}|/(|\tau_{mpc}|+|\tau_{pd}|)$ is also calculated, where $\tau_{pd}$ denotes the PD term in (\ref{eq:PD}). As expected, the torque command is faithfully executed, and the feedforward $\tau_{mpc}$ contributes the majority of the total torque.

\subsection{Scenario \#3: Environment Interaction}
\begin{figure}[!t]
        \vspace{-0.3cm}
        \centering
        \subfloat{\includegraphics[width=.48\textwidth]{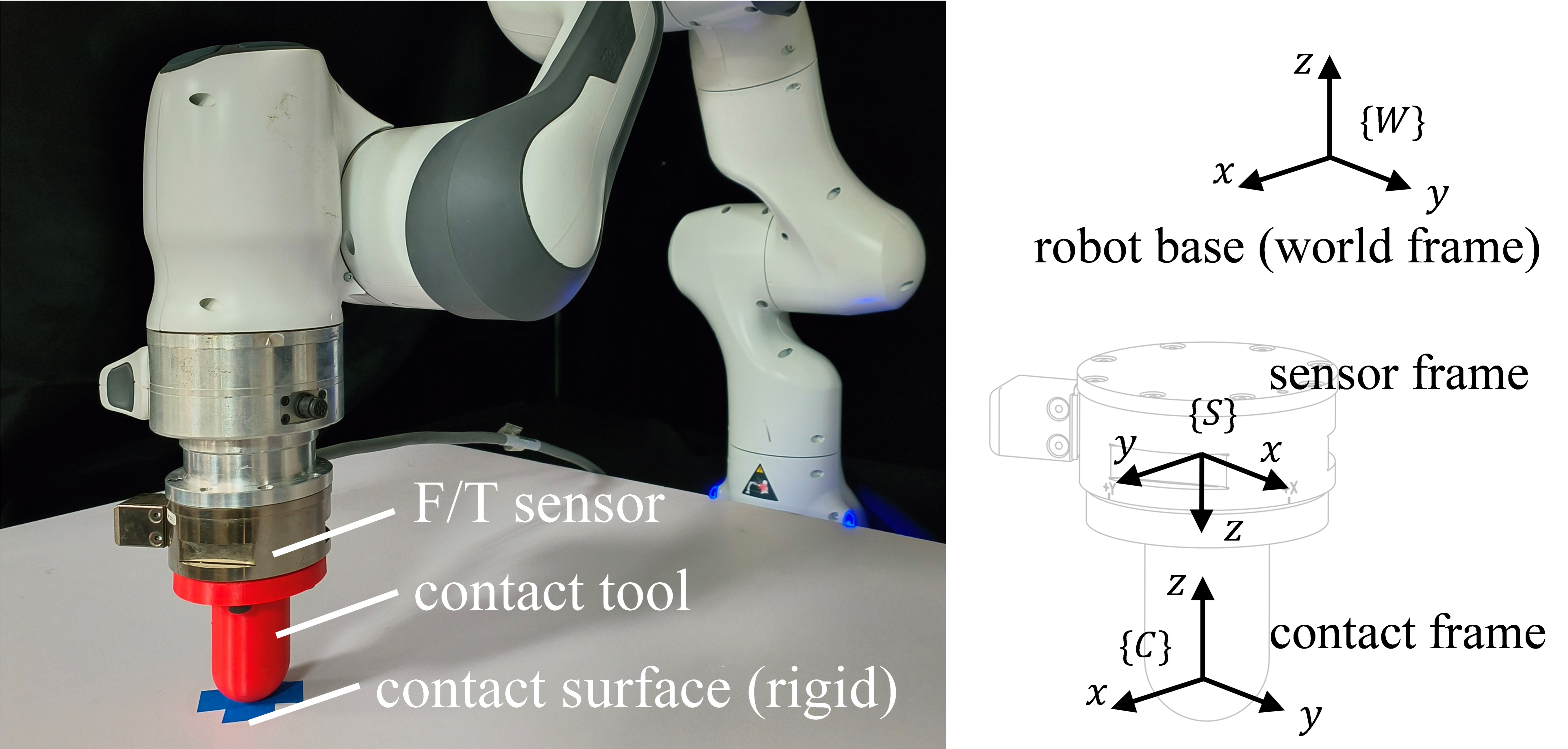}} \\
        \caption[Scenario-3]{Scenario \#3: Evaluation of environment interaction. \textit{Left:} Experimental setup. \textit{Right:} Sensor frame and contact frame relationship. Notably, the F/T sensor is solely used for evaluation, without being involved in feedback control.}
        \label{fig:3_exp}
    \end{figure}

\begin{figure}[!t]
        \centering
        \subfloat{\includegraphics[width=.48\textwidth]{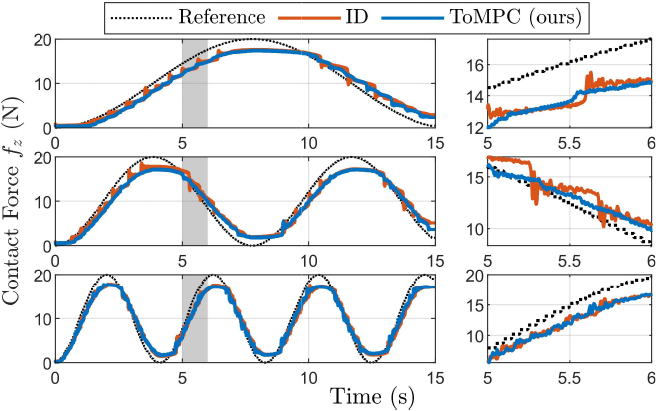}} \\
        \vspace{-0.2cm}
        \caption[Scenario-3-result]{Force measurement results for Scenario \#3. The diagrams from top to bottom correspond to the frequencies $w \!=\! 0.4, 0.8, 1.5$, respectively. The left grey-shaded area is magnified in the right diagram. }
        \label{fig:3_exp_results}
        \vspace{-0.4cm}
    \end{figure}

    The environment interaction performance of the ToMPC is evaluated in the third scenario. To measure the contact force, a ATI Gamma F/T sensor is mounted, and the robot gripper is replaced with a customized contact tool as the end-effector, see Fig. \ref{fig:3_exp}. Notably, the sensor is used solely for evaluation purposes, without being involved in feedback control. A contact frame is defined at the robot tip, aligned with the world frame in orientation. In the experiment, we command the force reference trajectory $ \bm{\mathcal{F}}^{ref} \!=\! (0,\! 0,\! f^{ref}_z,\! 0,\! 0,\! 0)$ where $f^{ref}_z \!=\! 10\sin(wt\!-\!1.57) \!+\! 10$ N. The force-direction weight in $\bm{Q}_f$ is activated whereas the corresponding weight in $\bm{Q}_m$ is set to zero. The resulting measurement $f_z$ expressed in the contact frame is plotted in Fig. \ref{fig:3_exp_results}. For the three frequencies $w \!=\! 0.4, 0.8,1.5$, the ToMPC demonstrates a good alignment with the reference trajectory, validating the effectiveness of the interaction model (\ref{eq:dyn_env}) for force planning.
    Additionally, the ToMPC is compared to an instantaneous controller, which directly integrates the inverse dynamics (\textbf{ID}) (\ref{eq:dyn_robot}) with the interaction model. As shown in Fig. \ref{fig:3_exp_results}, the benchmark method exhibits oscillations during contact with a rigid surface, while the ToMPC achieves smooth trajectories with minimal noise. This damping of forces originates from the long-horizon dynamics optimization in the DDP. It enables the prediction of future behavior over a period of time and computes minimal control inputs, thereby filtering out noise and making the ToMPC suitable for high-fidelity force manipulation tasks.
    
\subsection{Scenario \#4: Hybrid Motion-Force Task}
    Based on the findings from Scenario \#2 and Scenario \#3, the superiority of ToMPC in hybrid motion-force tasks is highlighted in Scenario \#4. In this scenario, the robot is tasked with wiping a whiteboard along a circular trajectory in the X-Y plane while keeping a contact force of $8$ N along the Z-axis of the world frame. Similar to Scenario \#2, a ball obstacle exists in the workspace, and the oaMPC serves as the benchmark. Experimental snapshots of the task execution are shown in Fig. \ref{fig:4_exp}, and the complete process is available in the supplementary video.
    It can be seen that both methods yield collision-free motion during interaction with the environment. In this process, contact dynamics constraints are implicitly enforced through DDP shooting, and obstacle avoidance in kinematics is handled by the QP. The consistency error between the two optimization subproblems is mitigated via Lagrange multiplier updates. However, with regard to manipulation quality, the benchmark oaMPC is only capable of cleaning the first marker on the whiteboard.
    In comparison, benefited from proactive link avoidance (indicated by a green arrow), the ToMPC can explore a larger portion of the Cartesian space, successfully reaching two marker points. This result implies that the proposed task-oriented avoidance enhances manipulation efficiency in cluttered environments. Table \ref{table:jerk} gives quantitative metrics. The position error norm of ToMPC is much lower, though the force tracking error is slightly larger. Since the error amplitude remains within a small range, the trade-off in force accuracy is acceptable.

    \begin{figure}[!ht]
        \vspace{-0.3cm}
        \centering
        \subfloat{\includegraphics[width=.24\textwidth]{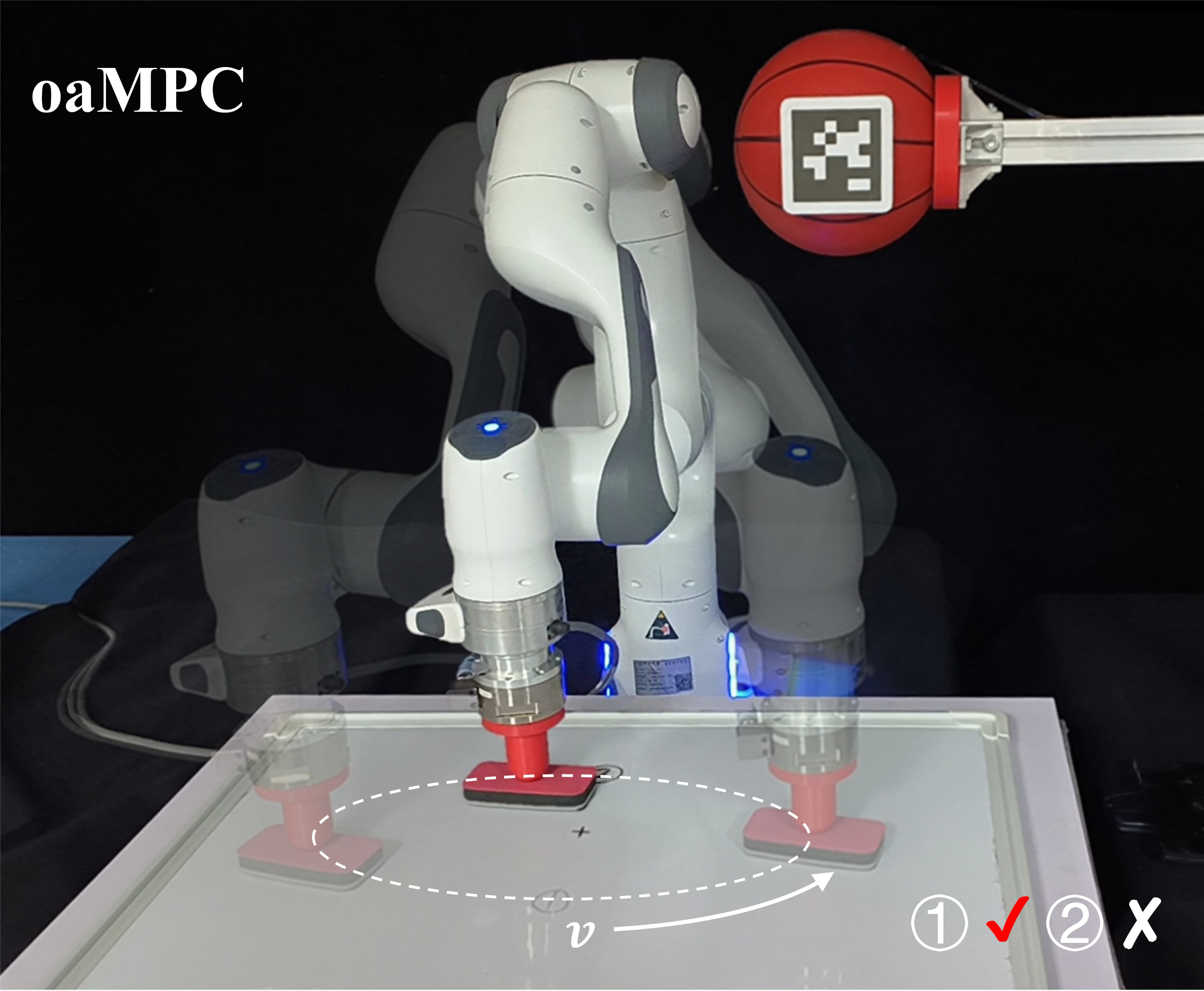}}
        \hfill
        \subfloat{\includegraphics[width=.24\textwidth]{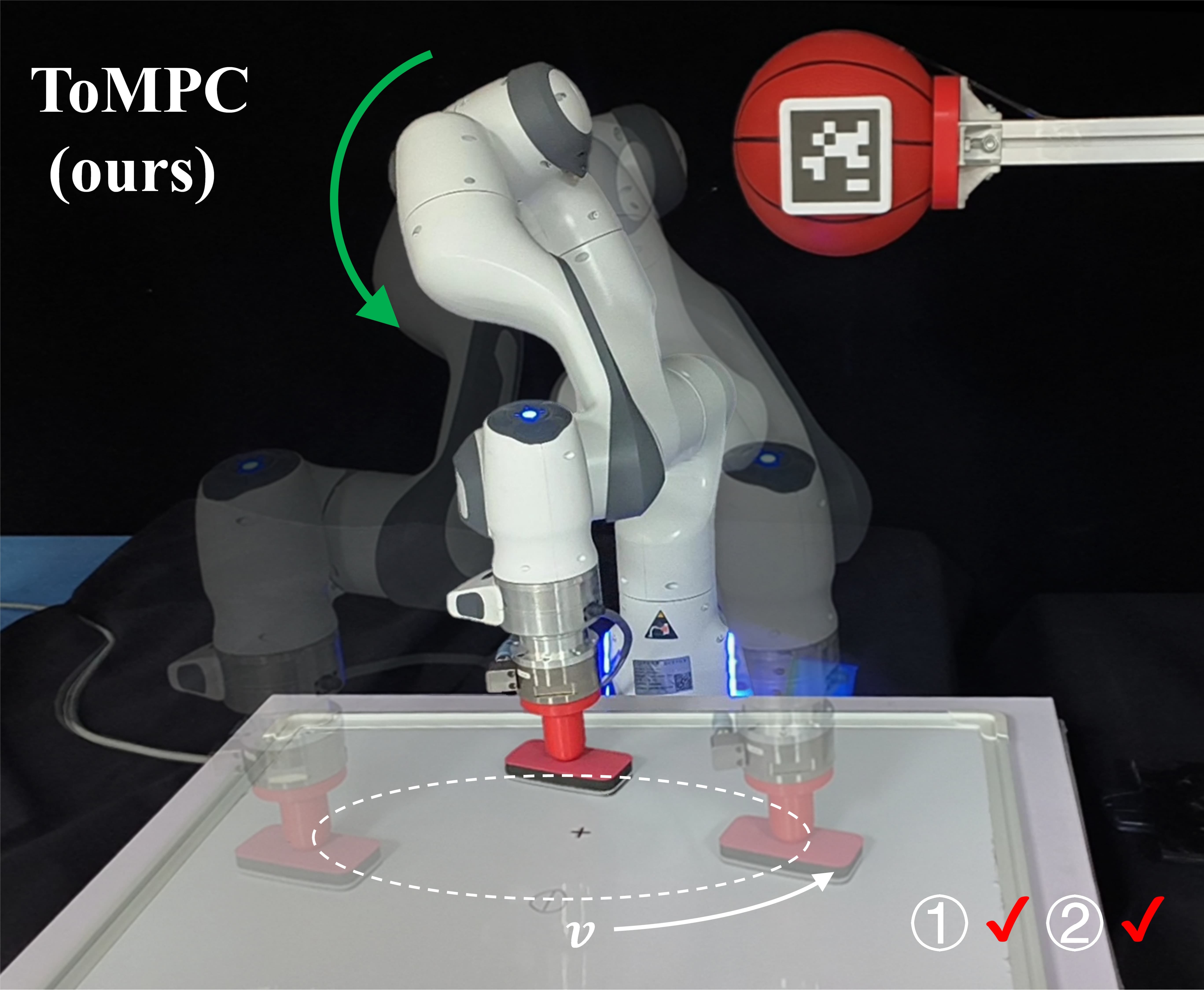}} \\
        \subfloat{\includegraphics[width=.24\textwidth]{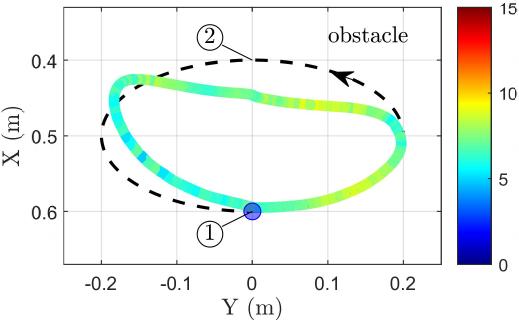}} 
        \hfill
        \subfloat{\includegraphics[width=.24\textwidth]{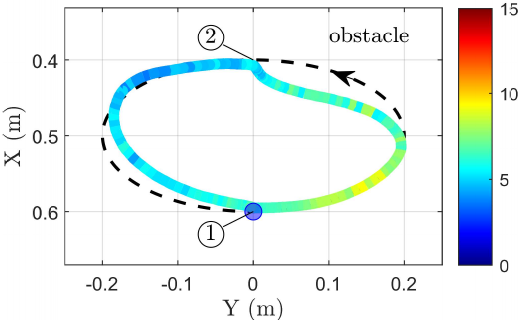}} \\
        \caption[]{Scenario \#4: The robot wipes a whiteboard with a contact force of $8$ N. \textit{Top:} Experimental snapshots of the oaMPC and the ToMPC methods. \textit{Bottom:} Rainbow plots illustrating the Cartesian trajectory and measured contact force.}
        \label{fig:4_exp}
        \vspace{-0.2cm}
    \end{figure}

\section{Conclusion}
\label{conclusion}
    \begin{table}[!t]
        \centering
        \caption{Comparison of Position and Force Errors in Scenario \#4} 
        \label{table:jerk}
        \begin{threeparttable}
            \renewcommand{\arraystretch}{1.2} 
            \begin{tabular}{c c c c c}
                \toprule
                \multirow{2}*{\textbf{Method}} & \multicolumn{2}{c}{\textbf{Position Error Norm} } & \multicolumn{2}{c}{\textbf{Force Error Norm}} 
                \\ \cmidrule(lr){2-5}
                & Mean (m) & RMS (m) & Mean (N) & RMS (N) \\
                \hline
                oaMPC   & $0.0303$  & $0.0355$ & $\bf{1.0505}$  & $\bf{1.2425}$ \\
                ToMPC   & $\bf{0.0201}$  & $\bf{0.0236}$ & $2.0335$  & $2.3020$ \\
                \bottomrule
            \end{tabular}
        \end{threeparttable}
        \vspace{-0.3cm}
    \end{table}
    
In this work, we propose the ToMPC planner, based on the ADMM, for safe and efficient manipulation in open workspaces. The approach plans collision-free motion and force trajectories for diverse manipulation scenarios within a unified MPC framework. The novel task-oriented avoidance strategy effectively increases the manipulation range in obstructed environments. Experimental results in simulation and on real hardware validate our idea. The proposed ToMPC enables real-time planning while strictly satisfying safety-related inequality constraints. However, the optimization performance in practice is sensitive to the quality of initial trajectories. Future work will explore learning-based methods for generating initial guesses and automatically tuning cost weights to achieve more robust deployment.

\bibliographystyle{IEEEtran}
\bibliography{references}

\end{document}